\documentclass[pdflatex,sn-mathphys-num,iicol]{sn-jnl}

\usepackage{graphicx}%
\usepackage{multirow}%
\usepackage{amsmath,amssymb,amsfonts}%
\usepackage{amsthm}%
\usepackage{mathrsfs}%
\usepackage[title]{appendix}%
\usepackage{textcomp}%
\usepackage{manyfoot}%
\usepackage{booktabs}%
\usepackage{algorithm}%
\usepackage{algorithmicx}%
\usepackage{algpseudocode}%
\usepackage{listings}%
\usepackage{musicography}
\usepackage[table,xcdraw]{xcolor}
\usepackage{arydshln}
\usepackage{subcaption} 
\usepackage{rotating}

\usepackage{fancyhdr}

\fancypagestyle{firstpage}{
    \fancyhf{}
    \fancyfoot[C]{\small Preprint. This manuscript is under review at Pattern Recognition Letters.}

}

\theoremstyle{thmstyleone}%
\theoremstyle{thmstyletwo}%

\theoremstyle{thmstylethree}%

\begin{document}
\renewcommand{\arraystretch}{1.1}

\title[Direct content-based retrieval from music scores images]{Direct content-based retrieval from music scores images
}

\author*[1]{\fnm{Noelia} \sur{Luna-Barahona}}\email{noelia.luna@ua.es}
\author[1]{\fnm{Antonio} \sur{Ríos-Vila}}\email{arios@dlsi.ua.es}
\author[1]{\fnm{Félix} \sur{Fuentes-Hurtado}}\email{ffuentes@dlsi.ua.es}
\author[1,2]{\fnm{David} \sur{Rizo}}\email{drizo@dlsi.ua.es}
\author[1]{\fnm{Jorge} \sur{Calvo-Zaragoza}}\email{jcalvo@dlsi.ua.es}

\affil[1]{\orgname{Pattern Recognition and Artificial Intelligence Group, University of Alicante}, \country{Spain}}
\affil[2]{\orgname{Instituto Superior de Enseñanzas Artísticas de la Comunidad Valenciana}, \country{Spain}}

\abstract{
The digitization of musical scores plays a crucial role in their preservation and accessibility, yet information retrieval still depends mainly on metadata searches, such as by title or composer. Content-based search in music score images remains underexplored compared to text documents, despite its potential value for musicians, musicologists, and educators. This work contributes to the field by first studying which characteristics of a score are most relevant for search and by defining a systematic method to build query datasets from any annotated corpus. We also consider diverse methods for content-based search on music score images, ranging from transcription-based approaches relying on Optical Music Recognition (OMR), to a transcription-free Transformer model trained to recognize queries directly from score images, and a text-prompted Large Language Model. Our experiments evaluate these models on four corpora exhibiting diverse characteristics in terms of dataset size, image quality, and typesetting mechanisms. Overall, each method excels under different conditions: OMR-based pipelines achieve higher in-domain retrieval, whereas transcription-free models handle domain variability more effectively. 
}




\maketitle

\thispagestyle{firstpage}

\section{Introduction}\label{introduction}
    Music plays a vital role in society, reflecting cultural and social transformations throughout history, and as such it deserves to be accessible to everyone and preserved with the same rigor as other forms of cultural heritage. The digitization of music is therefore of the foremost importance, both for ensuring the continuity of music knowledge---especially given that compositions have traditionally been documented in paper, which is highly susceptible to many forms of degradation---and facilitating broader public access to these works by making them openly available or enabling precise retrieval and filtering of relevant content.
    
    While music archives are increasingly digitized~\citep{kurth2008framework}, scanned images alone do not currently permit music information retrieval (MIR) or computational musicology tasks. For scores to achieve machine-interpretability, they are typically transcribed into symbolic encodings---a process that is labor-intensive when done manually, and that can be automated through Optical Music Recognition (OMR) systems~\citep{calvo2020understanding}. However, while OMR has been a longstanding topic of research in MIR, its results are far from perfect, which affects negatively downstream tasks. Consequently, tasks beyond transcription remain crucial for practical applications of musical data. This is the reason why our work focuses on content-based search over untranscribed scans of music scores, to overcome the limitations imposed by reliance on OMR.
    
    
    Content-based search refers to the retrieval of documents according to the information they contain~\citep{yoshitaka1999survey}, as opposed to approaches based on metadata, such as title, author or caption, depending on the file type~\citep{schedl2016lfm}. 
    Designing an effective content-based retrieval system for music scores can enable a variety of valuable applications. For instance, it could allow users to locate compositions based on a remembered fragment---similar to current capabilities in audio retrieval---or to identify recurring motifs across an artist's body of work, as well as to detect potential cases of melodic plagiarism~\citep{de2017music} and duplicate search~\citep{crawford_lute}. Beyond analytical and archival purposes, such systems also have pedagogical potential, such as identifying pieces that contain technically demanding passages for a given instrument~\citep{ramoneda2024combining}. These applications are widely relevant---supporting large-scale initiatives in the music industry and heritage preservation, while also addressing the practical needs of individual musicians and educators. As such, this research direction is highly relevant and provides a strong motivation for the present study.

    In the domain of natural language processing, content-based search within textual documents (also called full-text search) is a mature and extensively studied field~\citep{blair1985evaluation}. In contrast, its application to music scores remains limited, particularly on the case of untranscribed images~\citep{zhu2022framework}. Relevant works are covered in the Section~\ref{background}.
    
    Our contributions address multiple facets of content-based music retrieval. First, we present a study of existent content-based music queries, identifying a set of relevant query types for symbolic search over scores and organizing them according to their musical features and retrieval needs. Second, since no existing dataset is specifically designed for supervised training on content-based search over untranscribed music score images, we construct one from scratch---developing a methodical framework for generating musically meaningful queries from any dataset originally annotated for transcription, as these are readily available. Finally, we explore multiple methodologies for content-based retrieval from sheet music images, including traditional transcription-based pipelines, fine-tuning multimodal large language models, and a compact end-to-end architecture---the latter two of which eliminate the need for symbolic transcription. We conduct an exhaustive empirical evaluation of these approaches across varied datasets and experimental conditions, analyzing how well each method generalizes and how factors such as transcription errors affect retrieval performance.

    

\section{Background}\label{background}
    Typically, music (whether audio recordings, scanned images or symbolic scores) is retrieved based on metadata, such as the title of the piece or the name of the composer~\citep{bainbridge2003,hosey2019}. In contrast, music identification ---the process of recognizing a recorded performance from an audio excerpt---is an area of study that has seen significant advancements.
    
    For example, 
    the work of \citet{serra2009cross}, studies how to discriminate whether two recorded songs are covers of each other. 
    \citet{grosche_et_al:DFU.Vol3.11041.157} studies the retrieval of musical audios according to a given audio query, comparing different approaches for three different tasks: audio identification, audio matching and version identification. 
    \citet{salamon2013tonal} also tackles audio version identification, and adapts their methodology to also allow query-by-humming retrieval. 
    The more recent review by \citet{aly2024databasequerying} focuses on works that model and query databases of symbolic musical scores.
    These methodologies are centered on the musical content instead of relying on metadata. This raises the question of what would be the equivalent search method for musical score images.
    
    As of today, many of the content-based search methods for written music in the literature rely on the availability of symbolic scores for query comparison~\citep{typke2005survey,zhu2022framework}. Since scanned sheet music must be transcribed into symbolic format before such methods can be applied, content-based retrieval is inherently tied to the success of OMR systems. 

    \subsection{Optical Music Recognition}
        %
        
        
        
        
        Many music analysis and retrieval tasks rely on having access to a symbolic, machine-readable representation of notated music. Traditionally, obtaining these symbolic transcriptions has required expert human labor, often performed by musicologists or trained transcribers---a process that is both time-consuming and costly.
        
        Optical Music Recognition (OMR) aims to automate this task by converting scanned or photographed sheet music into symbolic formats such as MusicXML~\citep{good2001musicxml}, ABC~\citep{walshaw2011abc}, MEI~\citep{roland2002music} or kern~\citep{huron1997kern}.
        
        Early OMR systems relied heavily on extensive image preprocessing and classical object recognition techniques to detect staff lines, segment musical symbols, and transcribe them individually \citep{rebelo2012optical}. 
        With the rise of deep learning, neural-based approaches emerged, offering end-to-end transcription pipelines~\citep{calvo2018end} 
        that bypass much of the manual preprocessing. These models have advanced to the point where they can handle full-page score images in a single pass \citep{smt-RiosVila}.
        
        Despite these advances, a key limitation of current systems---both traditional and neural---is their sensitivity to domain shifts~\citep{luna2024unsupervised, smt-RiosVila, gold2018_ftempo}.

    \subsection{Content-based search systems for symbolic music scores}\label{queries}
        A contribution of this work consists of studying and assessing what constitutes efficient, interpretable, and musically relevant queries, with the purpose of applying these insights in our research. Consequently, the first step taken was to study in detail the types of queries used in previous studies, in order to establish a point of reference in a field that has not seen significant advancements in years.
        
        \citet{downie1999evaluating} proposes the use of n-grams, a natural language processing approach, to musical documents, 
        which allows the use of traditional text-based information retrieval techniques on musical documents.
        
        \citet{lemstrom1998musical} formulate queries as sequences of interval--duration pairs, integrating both pitch movement and rhythmic structure. 
        Unlike the approach proposed by \citet{downie1999evaluating}, which relies solely on pitch-based features, this method explicitly incorporates rhythm.
        
        In 2004, the \textit{Themefinder}\footnote{\url{http://www.themefinder.org}} database and web tool~\citep{sapp2004search} were released, enabling searches within kern-encoded musical scores using queries based on pitch (scale-invariant), interval, scale degree and contour
        , with no restriction on query length and allowing wildcard tokens for advanced search options. It also supports searches based on meter and key. 
        This study has been instrumental in defining the query types we aim to use in our work.
        
        \textit{Musipedia}\footnote{\url{https://www.musipedia.org/}}~\citep{doi:10.1177/102986490901300109} is another example of a database and online tool for content-based music search. It supports queries based on melody (notes and durations), contour, humming, and rhythm tapping. 
        While \textit{Themefinder} assumes familiarity with modern musical notation, \textit{Musipedia} is designed to be more accessible to non-expert users. Query length is restricted to a range of 5 to 25 tokens, and it computes edit distance between the query and the searched documents to allow retrieval of approximate matches.
        
        Another comparison approach proposed by \citet{urbano2011melodic} represents the melody of a musical segment as a pitch-time curve and evaluates similarity by analyzing differences in the shape transformations between the query's curve and those of the database entries.
        
        The latter Musiconn score search engine\footnote{\url{https://scoresearch.musiconn.de/ScoreSearch/about?lang=en}}~\citep{Pulimootil2018} continues to follow the search-over-transcription paradigm. However, even in this modern system, the authors acknowledge that OMR-generated transcriptions are prone to errors, which can impact the reliability of downstream retrieval tasks.

        The work \citet{gold2018_ftempo} explores identifying duplicates and closely related pages in a music collection by grouping the tokens in their symbolic encodings (MEI mensural) into ``words'' using two approaches: the aforementioned n-grams, and minimal absent words. Pages are compared based on the number of words they have in common. The authors describe this method as resilient to OMR-related errors, but this search method is limited in that it only allows comparisons at full-page level. This methodology is used in F-TEMPO\footnote{\url{https://f-tempo.org/}}.
        
        The previous work is studied by \citet{crawford_lute}, exploring its potential as a tool for musicologists. They build on \citet{gold2018_ftempo}'s page-matching experiments and introduce a new approach. Using symbolic lute intabulations, they extract sequences of the highest notes at regular time intervals and compare them to the highest voice in vocal scores indexed in F-TEMPO. However, the exact composition of these queries—including their length and whether they correspond to full pieces or excerpts—is not clearly specified.

        In 2021, the \textit{Répertoire International des Sources Musicales} (RISM)\footnote{\url{https://rism.info/}} launched RISM Online\footnote{\url{https://rism.online/}}, a search engine for the RISM Catalog that enabled content-based queries by exact pitch, contour or intervals. This feature was possible because the scores included symbolically encoded musical incipits in MEI, as explained in \citet{McKay_2024}.
        
        In a more recent study, \citet{rigaux2024topological} propose Muster, a data model to represent a symbolic music score as graphs. This model relies on a structured representation of a music score's contents and enables querying over it. However, constructing this representation requires a prior symbolic transcription of the music score (in this case, in MEI notation.
        
        Although earlier studies laid the groundwork for content-based retrieval in symbolic music, much subsequent work has considered retrieval over uniformly encoded representations, where the task often becomes a standard retrieval problem. This reliance on accurate transcriptions, which cannot be guaranteed in practice, limits their applicability. In this work, we assess how some of these established querying strategies perform on automatically transcribed---and therefore potentially noisy---musical representations, as well as investigate alternative methods that operate directly on score images.

    \subsection{LLMs and music data}
        In recent years, Large Language Models (LLMs) have experienced a remarkable surge in popularity, driven by their substantial improvements in language understanding and generation. These models are now regarded as general-purpose systems, capable of addressing a wide range of tasks thanks to their broad linguistic and contextual reasoning abilities. With the introduction of multimodal architectures, LLMs can also process and interpret visual inputs, aligning image and text representations to perform tasks such as object recognition, scene understanding, and image description.

        Given their versatility, LLMs have become a benchmark in artificial intelligence research, where novel tasks are typically assessed on these models to determine whether they can be accomplished without explicit domain-specific training.

        In the domain of music, LLMs demonstrate a strong grasp of theoretical and historical knowledge \citep{calvo2024can}. They can discuss harmony, form, and style with notable accuracy, as well as reference composers, historical periods, and major works---reflecting the wealth of textual information about music available on their training corpora. However, this competence remains primarily textual and declarative: while LLMs can articulate what musical concepts are and describe stylistic or historical contexts, they struggle to apply this knowledge to actual musical materials. Tasks that involve reading and interpreting music scores, which require translating visual notation into symbolic or auditory understanding, continue to pose a significant challenge \citep{calvo2024can}. This gap highlights the current limitations of LLMs in bridging theoretical and historical context with perceptual and symbolic interpretation \citep{cuskley2024limitations}.


        A few recent preliminary studies have begun exploring the application of LLMs to music recognition, but their results remain limited: some report substantial retrieval errors~\citep{tang2025nota}, while others evaluate only on the same synthetic data used for training~\citep{chen2025musixqa}. Although these works represent important first steps, their setups either perform poorly on the task or fail to reflect real-world conditions.
        
\section{Search queries in sheet music}\label{search}
    
    As previously discussed, content-based search in musical scores has a wide range of potential applications, from analyzing a piece's technical difficulty~\citep{ramoneda2024combining} to identifying compositions that share similar musical motifs~\citep{hsiao2023bps}. Consequently, a significant part of this work has focused on studying the characteristics of musical scores that may be valuable to retrieve in various contexts~\citep{hewlett2004music}. This work represents a first, foundational step in our investigation, serving as an essential proof of concept. By focusing initially on basic, objectively measurable queries, we are able to establish a clear baseline that allows meaningful comparisons between symbolic and image-based search methods. This foundational step is necessary to validate our approach and ensure that subsequent experiments with more elaborate queries or partial-matching strategies can be interpreted rigorously and reliably.
    
    Since no existing dataset is designed for supervised training in this task, we constructed one from scratch---developing a systematic framework for generating musically meaningful queries from any dataset annotated for transcription. This framework 
    enables consistent evaluation of retrieval methods across diverse musical corpora.
    
    
    
    As we designed it, there are two distinct types of queries, and each of them will follow a slightly different approach. On the one hand we have the queries that refer to a single symbol, which can be or not be present in the score. On the other hand are the queries of sequences: they refer to multiple consecutive symbols, such as a melody, and have to be all present and ordered for the query to be considered correct.
    
\begin{figure*}
        \centering
        \includegraphics[width=0.55\linewidth]{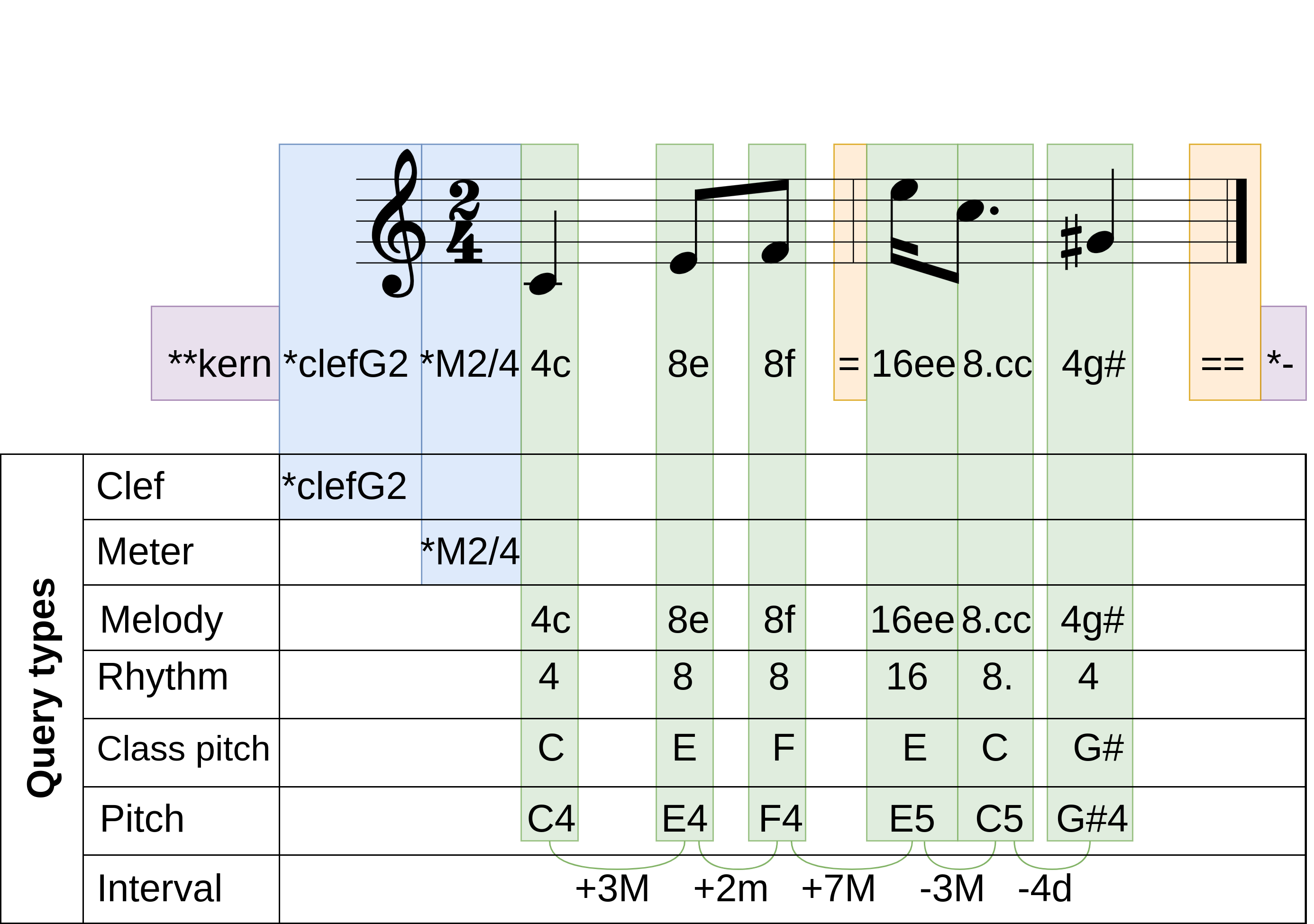}
        \caption{Representation of the different queries considered over an example staff. Under the staff we find the equivalent kern notation. Neither the elements specifics to the kern-notation nor the staff separation bars are used in any query information.}
        \label{fig:queries-example}
    \end{figure*}

    The final queries selected for this work, as well as their justifications, can be found in what follows (examples given in Figure~\ref{fig:queries-example}).
    

    \textbf{Clef.} Whether a musical clef is present in the fragment. Musical clefs are represented by a single symbol in the graphical music score, and as a single token in kern notation. Identifying clefs in a score is valuable, because clef usage is intrinsically tied to an instrument's pitch range to keep notation within the staff and avoid excessive ledger lines, improving readability (see the \textit{Notation of Notes, Clefs and Ledger Lines} in \citet{gotham2021openmusic}). We consequently believe retrieving clef information could enhance document search or filtering, providing further information to determine the intended instrument or vocal part. 
    
    
    \textbf{Meter.} In kern notation, meter indications are considered as a single token. Time signatures in music scores define the rhythmic structure and pulse of the piece, indicating how beats are grouped and accented (\textit{Notating Rhythm}, \citet{gotham2021openmusic}). From a practical perspective, retrieving meter information can benefit filtering or grouping scores, allowing users to focus on pieces with specific rhythmic characteristics, facilitating comparative studies or targeted learning.
    
    To facilitate their search, both the clef and meter queries are encoded in kern format.
    

    \textbf{Melody.} Melody-based queries combine both pitch and rhythmic information. This dual encoding enables precise matching of complete musical phrases or fragments, making it especially useful to retrieve exact or closely related passages. Unlike isolated pitch, rhythm, or interval queries, melodic queries capture the full linear shape of a passage, allowing for fine-grained comparisons and reducing ambiguity in retrieval results. This makes them particularly effective for applications such as thematic indexing, motif identification, and musical similarity search within score databases.
    
    Once again, to simplify the comparison, we use the kern encoding (simplified to only the pitch and rhythm characters).
    
    
    \textbf{Rhythm.} Rhythm-based queries focus on the durational structure of musical notes. This approach is particularly useful for identifying and retrieving rhythmically complex or distinctive passages, such as syncopations, tuplets or irregular groupings. In educational and pedagogical contexts, rhythm queries can support the discovery of challenging exercises, providing a flexible way to organize and access repertoire based on difficulty, style, or instructional focus.
    
    Note durations are represented numerically, with the whole note (\musWhole) as the base unit valued at 1. Other notes are expressed as fractions of this unit, as is standard in western music notation~\citep{gotham2021openmusic}.
    
    
    \textbf{Pitch.} This query defines a sequence of consecutive pitches, ignoring rests and bar separators. From an information retrieval perspective, indexing and searching scores based on pitch content allows users to identify thematic material, compare melodic sequences, or detect motifs across documents. This is particularly useful in applications like plagiarism detection, music recommendation, or thematic cataloging.
    
    We differentiate between exact pitch and pitch class queries. Exact pitch queries are encoded in scientific pitch notation (e.g., C4), making them suitable for tasks requiring precise melodic or harmonic matching. In contrast, pitch class queries disregard the octave, allowing users to search for notes based solely on their pitch name (e.g., any C, regardless of octave). By offering both options, the system can serve a wider range of musical analysis and retrieval scenarios---from detailed forensic tasks to more abstract, register-independent pattern searches. 
    These queries are encoded in scientific pitch notation.
    
    
    
    
    \textbf{Interval.} Intervals are the differences in pitch between two notes~\citep{schachter2011harmony}. This is why we offer interval-based queries as an alternative, pitch-independent representation of melodies, similar to countour queries in other works~\citep{sapp2004search,phanplainchant}. By encoding the relative distance between successive pitches, interval queries enable the detection of transported patterns. These are highly effective for identifying recurring motifs, melodic contours, and structural relationships across different keys and registers, supporting robust pattern-based retrieval and analysis. 
    
    

    The encodings for each query type have mainly been chosen because they are the unprocessed output of the Humdrum tools package, which allowed for an easier construction and possible future extension of the queries dataset.
    
    Contour, humming or rhythm-tapping queries have been discarded for this work. These query-types are perfect to look for audio excerpts, but they rely on approximate matching. In this preliminary work, we validate the methodology using deterministic tasks, with the framework designed to be readily adaptable to diverse query types in future work.
    

\section{Approaches}\label{approaches}


    This section outlines existing and proposed methodologies that are considered suitable for addressing the query-matching task, including a traditional OMR-based approach with subsequent symbolic search and two models that bypass transcription altogether. Their specific implementation will be detailed on Section~\ref{exp-setup}.

    \subsection{Transcription-based matching}

\begin{figure}
            \begin{subfigure}[b]{\linewidth}
            \centering
            \includegraphics[width=\linewidth]{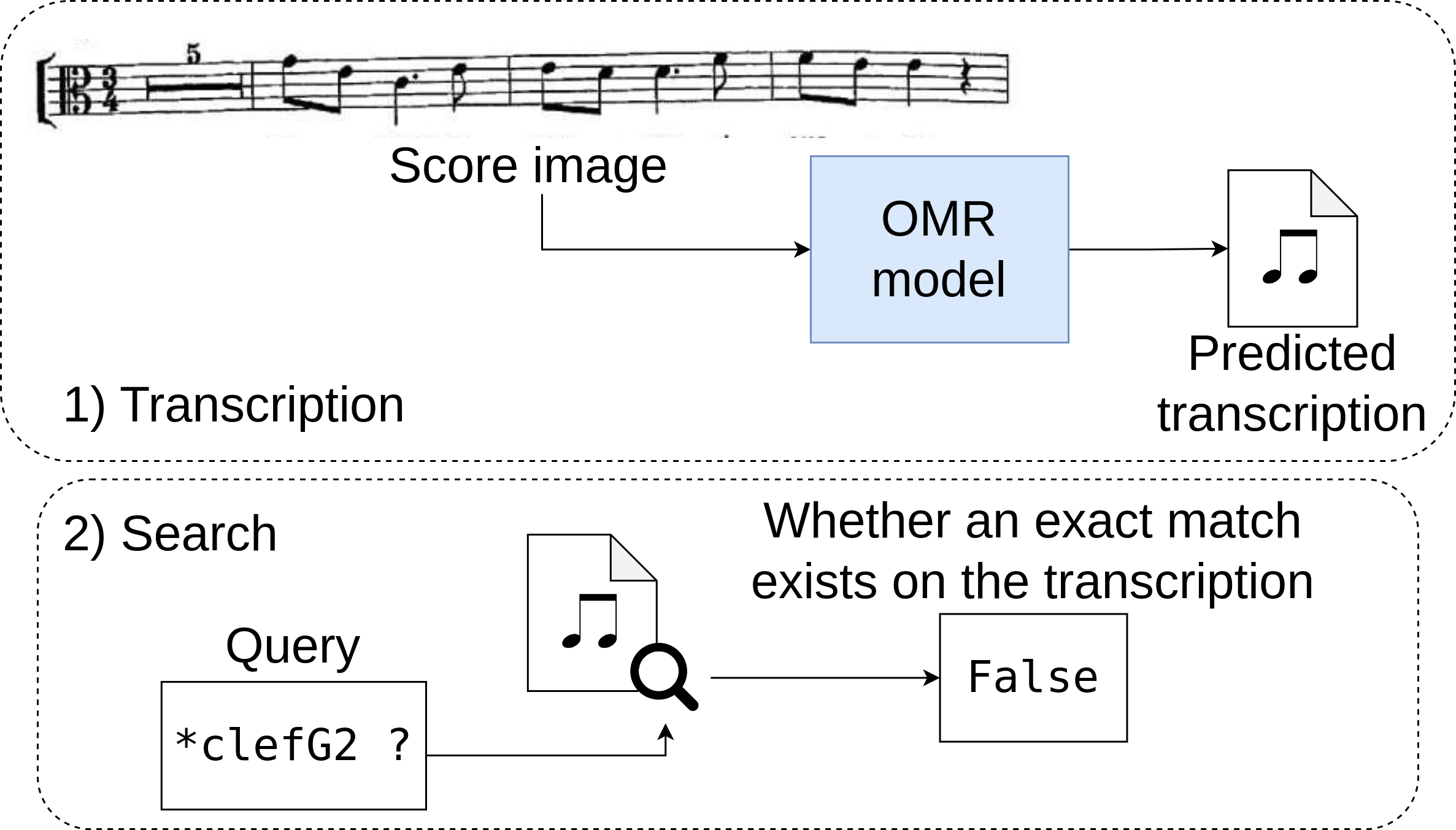}
            \caption{Transcription-based pipeline, where the OMR step and the content search are completely separate, but the retrieval is dependent on the transcription results.}
            \label{fig:appr_tr}
            \end{subfigure}
            \hfill
            
            \begin{subfigure}[b]{\linewidth}
            \centering
            \includegraphics[width=\linewidth]{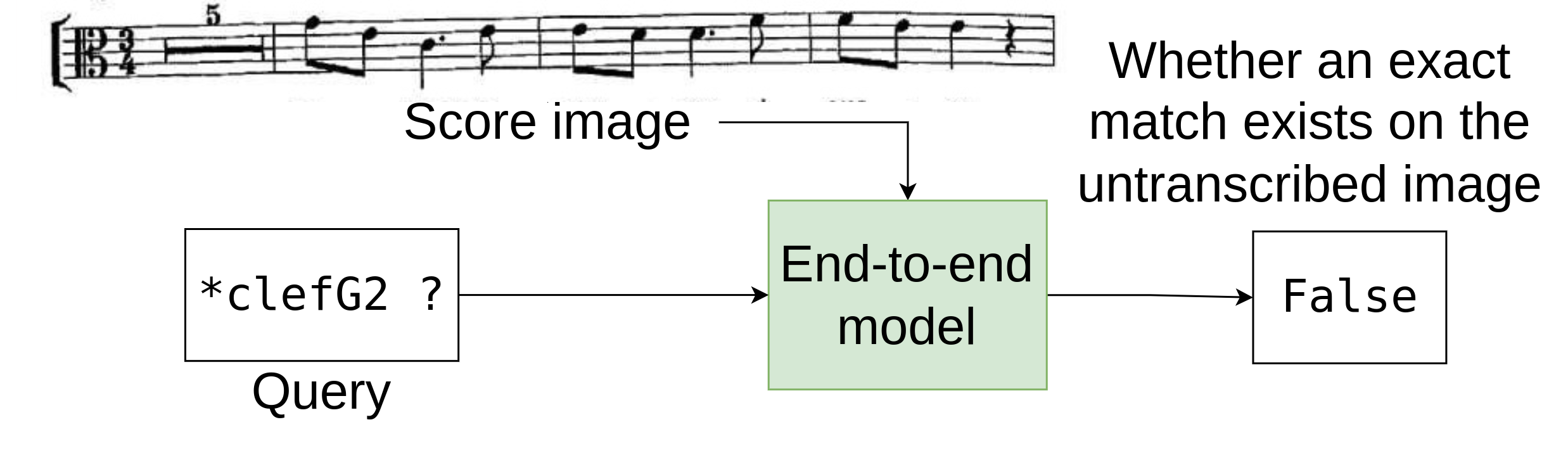}
            \caption{End-to-end pipeline, where a model is provided a staff image and a query, and must classify whether it is a match.}
            \label{fig:appr_e2e}
            \end{subfigure}
            \hfill
            
            \begin{subfigure}[b]{\linewidth}
            \centering
            \includegraphics[width=\linewidth]{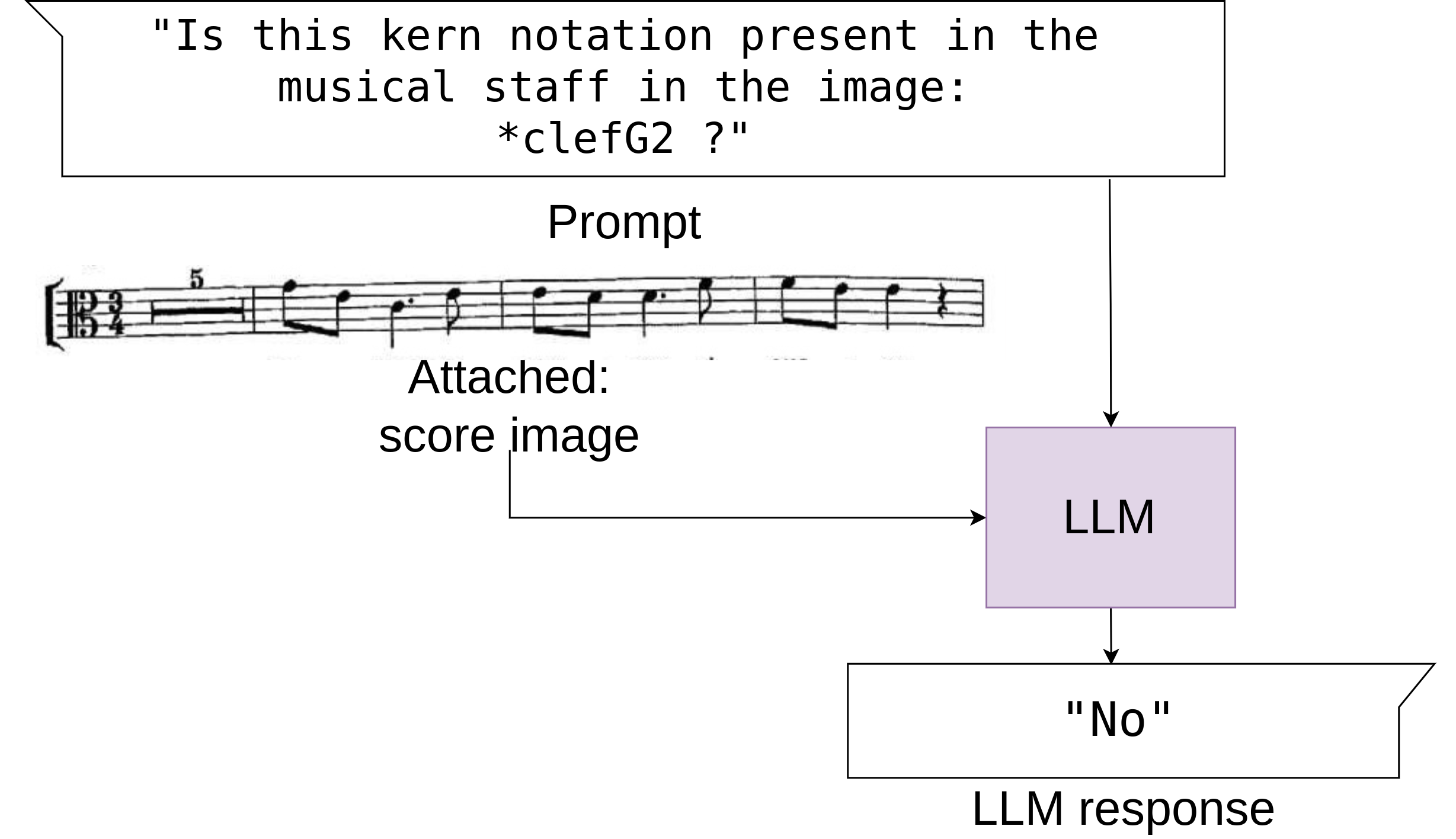}
            \caption{An LLM is fed a prompt and an attached staff image, and must answer the question from the prompt.}
            \label{fig:appr_llm}
            \end{subfigure}
        
        \caption{Diagrams of each of the three approaches for content-based search studied.}
        \label{fig:approaches}
        \end{figure}

        In this first set of experiments, we adopt the traditional pipeline for OMR \citep{calvo2020understanding} to first transcribe score images into symbolic notation before applying content-based retrieval techniques. This serves as our baseline approach (Figure~\ref{fig:appr_tr}).
        
        Once transcribed, symbolic files are converted into the query-specific encodings described in Section~\ref{search}. We employ an exact-matching strategy to locate occurrences of each query in the converted transcription sequences. Our objective is to evaluate how reliably each type of query can be recovered, highlighting the impact of OMR accuracy on retrieval performance.

    \subsection{End-to-end information retrieval}

        This approach represents an alternative to the traditional OMR pipeline by removing the transcription step entirely: the system directly determines whether a given musical query is present within the image (Figure~\ref{fig:appr_e2e}). 
        The retrieval task is therefore reframed as a binary classification problem over image-query pairs labeled as positive if the query is present anywhere on the image or negative otherwise. This allows the system to operate directly on raw visual data and avoids cascading errors from imperfect transcription.
        
        
    \subsection{LLM evaluation}
        Our study evaluates a fine-tuned multimodal LLM with image-parsing capabilities by presenting it with images of music scores and prompting it with 
        queries (Figure~\ref{fig:appr_llm}). 
        By analyzing its responses, we can determine whether the model is capable of extracting relevant information from the score and applying its theoretical knowledge to answer music notation queries.

\section{Experimental setup}\label{exp-setup}
    In this section, we describe the datasets, implementation details and configurations of the different approaches explored in this study.
    
    \subsection{Datasets}
        This work has been evaluated on two real modern sheet music corpora, and also makes use of a synthetic dataset to assess the utility of training with massive, out-of-domain data.

\begin{figure}[b]
            \begin{subfigure}[b]{\linewidth}
            \centering
            \includegraphics[width=\linewidth]{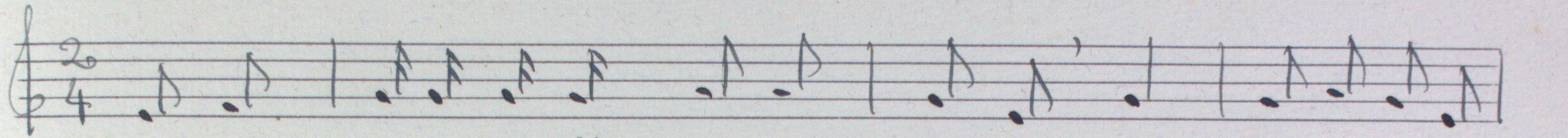}
            \caption{FMT-C}
            \label{fig:ds-fmtc}
            \end{subfigure}
            
            \begin{subfigure}[b]{\linewidth}
            \centering
            \includegraphics[width=\linewidth]{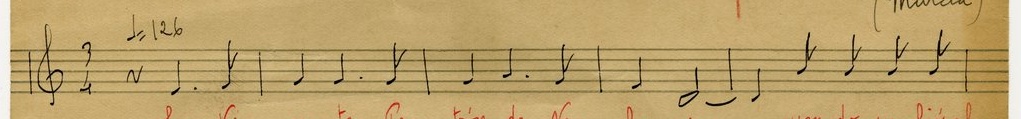}
            \caption{FMT-M}
            \label{fig:ds-fmtm}
            \end{subfigure}
            
            \begin{subfigure}[b]{\linewidth}
            \centering
            \includegraphics[width=\linewidth]{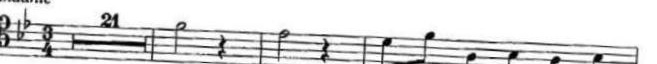}
            \caption{Malaga}
            \label{fig:ds-malaga}
            \end{subfigure}
            
            \begin{subfigure}[b]{\linewidth}
            \centering
            \includegraphics[width=0.5\linewidth]{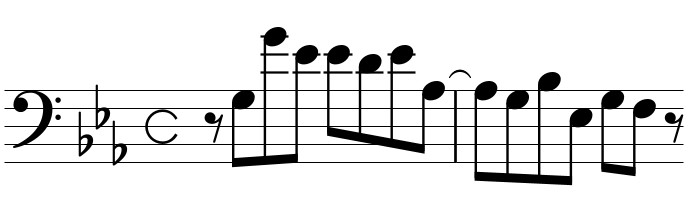}
            \caption{PrIMuS}
            \label{fig:ds-primus}
            \end{subfigure}
         \caption{Image samples from each of the datasets used in this work.}
         \label{fig:datasets-imgs}
        \vspace{-20pt}
        \end{figure}

        The ``Fondo de Música Tradicional'' corpus (FMT)~\citep{ros2021codified} is formed by traditional Spanish songs transcribed by hand. It contains scores with different origins and characteristics, so it is usually separated on two different clusters: \textbf{FMT-C} (Figure~\ref{fig:ds-fmtc}) and \textbf{FMT-M} (Figure~\ref{fig:ds-fmtm}). These partitions have clearly distinct visual features, such as the way the figures are drawn, the paper coloration or the thickness of the pen used.

        Traditional folk music typically features simple, singable melodies for broad participation and is often written within a standard vocal range using the G clef. Accordingly, all pieces in the FMT partitions are notated in the G clef. Clef-related queries were therefore omitted from these partitions, as the uniform notation prevents the model from learning meaningful distinctions.
        
        The ``Andalusian Music Documentation Center'' corpus (\textbf{Malaga})~\citep{Madueño:DLfM:2021} contains scanned, typeset music scores. Although such scores are usually easier for computer vision systems to process due to their uniform symbols and spacing, this dataset poses challenges because some staves are irregularly cropped. Figure~\ref{fig:ds-malaga} shows a specifically selected example of this phenomena.

        The \textbf{MultiDomain} dataset combines the previously described real datasets, enabling the study of whether increasing the number of real samples enhances generalization or, conversely, makes the learning process more difficult.

        Finally, an extract of the \textbf{PrIMuS} dataset \citep{calvo2018end} (Figure~\ref{fig:ds-primus}) was selected for the comparatively huge number of samples it provides (as we can see on Table~\ref{tab:datasets}), on account of being easier to obtain synthetic samples than to annotate real life corpora.
        
        \begin{table}[bp]
        \centering
        \begin{tabular}{l|lrr}
        \hline
        \textbf{Dataset} & \textbf{Format} & \textbf{N. pages} & \textbf{N. staves} \\ \hline
        FMT-C & Handwritten & 81 & 207 \\
        FMT-M & Handwritten & 161 & 722 \\
        Malaga & Typeset & 72 & 404 \\
        PrIMuS & Synthetic & - & 12395 \\
        \hline
        \end{tabular}
        \caption{Summary of the datasets employed, detailing their format and number of annotated pages and staves.}
        \label{tab:datasets}
        \vspace{-20pt}
        \end{table}
        

        All datasets include paired images of staves and their semantic encodings in the kern format, whose compactness reduces structural errors and aids neural network learning for transcription~\citep{rios2024end}.

        From each of the described corpus, we build a corresponding queries dataset. Positive queries are generated differently depending on the query type: for single-symbol queries, we extract all time signature and clef tokens present in each staff; for sequence queries, we convert the kern transcriptions into the appropriate encodings (defined in Section~\ref{search}) through Humdrum tools, and then extract substrings to serve as guaranteed positive query instances. Negative queries for each staff are created: (1) by selecting three examples that are positive for other staves but not for the current one, and (2) by modifying two positive queries to introduce 1–3 incorrect tokens, producing near-miss negatives. Consequently, each staff is associated with up to three positive and five negative queries per class.\footnote{The total number of queries of each type for every dataset partition is provided in the supplementary materials.} 

    \subsection{Implementation}

        This subsection describes the three model architectures considered in this work, along with their main design choices. We next detail each architecture individually.
        
        \subsubsection{CRNN for transcription}
             The transcription stage is performed using a Convolutional Recurrent Neural Network with Connectionist Temporal Classification (CTC-CRNN), a widely used architecture for sequence recognition tasks such as OMR~\citep{calvo2018end}. This model combines convolutional layers for visual feature extraction with bidirectional recurrent layers to model temporal dependencies across the horizontal axis of the score. The CTC loss function~\citep{graves2006connectionist} allows the network to learn flexible alignments between input images and output symbol sequences, making it suitable for recognizing staff-level annotations.
             
             Our implementation has 4 convolutional blocks with batch normalization, ReLU activations and MaxPooling, followed by 2 bidirectional LSTM layers with 256 units each, with dropout and lineal activation, as described by \citet{rosello2024source}. The model outputs sequences encoded in kern format, which is then used for exact query matching.

         \subsubsection{End-to-end Transformer}
             For this approach we use a Transformer architecture ---designed to process input-context pairs--- based on the Sheet Music Transformer network (SMT) proposed by \citet{smt-RiosVila}. The SMT integrates a ConvNext encoder to extract visual features from the score, and a Transformer decoder adjusts these features through cross-attention, conditioned on the query provided as context. The output projection of the decoder is replaced with a two-dimensional linear layer, enabling binary classification of the query-conditioned score representations into false and true classes.


             As our experiments target staff-level transcription---which require a less powerful and computationally-intensive network---we replaced the ConvNeXt of the original work with a simpler, more efficient convolutional neural network like the one described in the previous section, reducing training overhead. 
             
             Due to the novelty of the end-to-end approach and the lack of established optimal configurations, we perform a systematic exploration of architectural parameters by varying the convolutional filter sizes (32, 64, 128, and 256) and the number of decoder layers (1, 2, 4, and 8), in order to assess their impact on performance.
            
            We also considered leveraging the strong transcription performance of the model by using it as a pretrained backbone, then modifying its architecture and re-training on the binary classification task. However, due to the mismatch between the kern encoding tokens and those used in the query representations, this pretraining strategy did not yield any improvement. Consequently, these experiments were discarded.

         \subsubsection{Multimodal LLM}
             To evaluate whether a LLM could outperform a smaller Transformer model, we selected the PaliGemma 2~\citep{steiner2024paligemma} architecture, using its smaller 3-billion-parameter variant. This choice was motivated by the model's demonstrated effectiveness on the OMR task, where it achieved strong results. The best reported performance was obtained using an input resolution of 896px\textsuperscript{2}. However, due to computational constraints, we were unable to reproduce those experiments and instead employed a reduced resolution of 448px\textsuperscript{2}. We leave reproducing the experiments with the more suitable 896px\textsuperscript{2} resolution --as described in the original paper-- for future work.


            After extensive parameter tuning, we selected a batch size of 1 and 4-bit quantization to fit the model within the memory limits of a single NVIDIA A100 40GB GPU (our computational budget). Performance-driven choices included a learning rate of 1e-4, 16-bit LoRA \citep{hu2022lora}, and retraining the vision tower to improve convergence and accuracy. Fine-tuning was performed over 100,000 steps, as initial experiments showed no gains from longer training, with the best validation checkpoint used for testing.

    \subsection{Experiments}
        
        We evaluate the pipeline on both in-domain (ID) and out-of-domain (OOD) scenarios. This later setup reflects real-world conditions, where models must handle unseen data reliably. Additionally, we explore a multi-domain (MD) training setup, combining all available real collections. 
        
    \subsection{Metrics}
        To enable fair comparison, all retrieval methods are evaluated using a consistent framework: determining whether a query appears in a score fragment--using F1 as the primary metric for retrieval evaluation since it is preferred over accuracy due to the dataset's imbalance. To summarize performance across all query types, both macro and micro F1 are reported, where macro F1 computes the average F1 over each query type and micro F1 weights them by their frequency.
        
        For the transcription-based pipeline, we also examine how the quality of the symbolic output impacts retrieval performance 
        using the Symbol Error Rate (SER), a standard OMR metric 
        similar to the Character Error Rate (CER) in natural language processing, but applied to the full encoding of each musical symbol.

\section{Results}\label{results}

\begin{table*}[tbp]
\centering
\resizebox{0.7\linewidth}{!}{
\begin{tabular}{lll|rrrrrrrr}
\hline
& & & \multicolumn{8}{c}{\textbf{Test dataset}} \\
& & & \multicolumn{2}{c}{\textbf{FMT-C}} & \multicolumn{2}{c}{\textbf{FMT-M}} & \multicolumn{2}{c}{\textbf{Malaga}} & \multicolumn{2}{c}{\textbf{PrIMuS}} \\
& & & F1\textsubscript{M} & F1\textsubscript{m} & F1\textsubscript{M} & F1\textsubscript{m} & F1\textsubscript{M} & F1\textsubscript{m} & F1\textsubscript{M} & F1\textsubscript{m} \\ \hline
\multirow{15}{*}{\begin{turn}{90}\textbf{Training dataset}\end{turn}}
& \multirow{3}{*}{\textbf{FMT-C}} & TR & \cellcolor[HTML]{EFEFEF}\underline{73.8} & \cellcolor[HTML]{EFEFEF}\underline{75.2} & 11.8 & 9.6 & 5.2 & 2.1 & 7.8 & 5.4 \\
& & E2E & \cellcolor[HTML]{EFEFEF}52.9 & \cellcolor[HTML]{EFEFEF}49.3 & \underline{49.1} & \underline{48.2} & \underline{49.1} & \underline{48.1} & \underline{34.8} & \underline{46.3} \\
& & LLM & \cellcolor[HTML]{EFEFEF}51.6 & \cellcolor[HTML]{EFEFEF}57.9 & 25.2 & 29.3 & 22.1 & 28.2 & 27.5 & 26.6 \\ \cmidrule(l){2-11}
& \multirow{3}{*}{\textbf{FMT-M}} & TR & 5.1 & 3.3 & \cellcolor[HTML]{EFEFEF}\underline{85.9} & \cellcolor[HTML]{EFEFEF}\underline{86.1} & 7.2 & 3.6 & 8.3 & 5.7 \\
& & E2E & \underline{51.2} & \underline{53.8} & \cellcolor[HTML]{EFEFEF}52.8 & \cellcolor[HTML]{EFEFEF}54.0 & \underline{42.8} & \underline{50.4} & \underline{42.1} & \underline{50.3} \\
& & LLM & 31.1 & 37.9 & \cellcolor[HTML]{EFEFEF}50.6 & \cellcolor[HTML]{EFEFEF}55.6 & 33.1 & 39.6 & 33.1 & 40.2 \\ \cmidrule(l){2-11}
& \multirow{3}{*}{\textbf{Malaga}} & TR & 2.9 & 2.2 & 10.0 & 7.8 & \cellcolor[HTML]{EFEFEF}\underline{51.6} & \cellcolor[HTML]{EFEFEF}\underline{49.9} & 9.9 & 7.2 \\
& & E2E & \underline{48.1} & \underline{52.1} & \underline{53.4} & \underline{54.6} & \cellcolor[HTML]{EFEFEF}45.4 & \cellcolor[HTML]{EFEFEF}54.2 & \underline{51.0} & \underline{52.6} \\
& & LLM & 34.3 & 45.4 & 32.3 & 41.0 & \cellcolor[HTML]{EFEFEF}46.1 & \cellcolor[HTML]{EFEFEF}47.7 & 37.5 & 39.8 \\ \cmidrule(l){2-11}
& \multirow{3}{*}{\textbf{PrIMuS}} & TR & 1.1 & 1.4 & 1.0 & 1.0 & 4.8 & 1.5 & \textbf{98.1}\cellcolor[HTML]{EFEFEF} & \textbf{97.7}\cellcolor[HTML]{EFEFEF} \\
& & E2E & \underline{51.6} & \underline{52.6} & \underline{50.1} & \underline{49.9} & \underline{43.2} & \underline{56.5} & 74.9\cellcolor[HTML]{EFEFEF} & 74.6\cellcolor[HTML]{EFEFEF} \\
& & LLM & 30.8 & 42.8 & 30.7 & 38.7 & 36.4 & 37.8 & 81.6\cellcolor[HTML]{EFEFEF} & 79.1\cellcolor[HTML]{EFEFEF} \\ \cmidrule(l){2-11}
& \multirow{3}{*}{\textbf{MultiDomain}} & TR & \cellcolor[HTML]{EFEFEF}\textbf{76.3} & \textbf{77.6}\cellcolor[HTML]{EFEFEF} & \textbf{88.5}\cellcolor[HTML]{EFEFEF} & \textbf{88.2}\cellcolor[HTML]{EFEFEF} & \textbf{59.3}\cellcolor[HTML]{EFEFEF} & \textbf{58.7}\cellcolor[HTML]{EFEFEF} & 11.7 & 7.9 \\
& & E2E & 51.6\cellcolor[HTML]{EFEFEF} & 53.3\cellcolor[HTML]{EFEFEF} & 53.1\cellcolor[HTML]{EFEFEF} & 53.3\cellcolor[HTML]{EFEFEF} & 44.2\cellcolor[HTML]{EFEFEF} & 54.2\cellcolor[HTML]{EFEFEF} & 47.7 & 54.9 \\
& & LLM & 45.9\cellcolor[HTML]{EFEFEF} & 50.7\cellcolor[HTML]{EFEFEF} & 43.3\cellcolor[HTML]{EFEFEF} & 47.8\cellcolor[HTML]{EFEFEF} & 30.1\cellcolor[HTML]{EFEFEF} & 33.1\cellcolor[HTML]{EFEFEF} & \underline{68.7} & \underline{65.2} \\
\hline
\end{tabular}
}
\caption{Overview of the performance of the three search approaches across training-test dataset pairs. Macro (F1\textsubscript{M}) and micro (F1\textsubscript{m}) F1 scores are reported, with in-domain cases highlighted in gray. For each training--test pair, the highest scores are underlined, while the overall top scores for each test dataset are highlighted in bold.}
\label{tab:overview}
\end{table*}

    Since this work involved extensive experimentation, we focus on the results from the best configuration for each setting. We begin with an overview presenting the performance of the three different search approaches studied. Then, we will analyze each approach in more detail.

    From Table~\ref{tab:overview}, several observations can be made. First, as expected, ID experiments consistently outperform OOD ones across all datasets and approaches. Moreover, MD training proves particularly beneficial for the transcription task, yielding better results on real datasets as the model learns to generalize more effectively. 
    In contrast, the end-to-end, multi-domain training yields no improvement, and the LLM approach performs considerably worse with this training.

    Our transcription experiments yield markedly lower F1 scores compared to other approaches when evaluated on OOD test data. This arises when the transcription obtains high SER values, indicating poor alignment with the target. In such cases, nearly all positive queries fail to match the random text. In contrast, a random binary classifier applied to our unbalanced dataset, with approximately 40\% positive instances per class, would achieve a F1 score of approximately 0.44.\footnote{To obtain a further clarification of this point, the reader is referred to the supplementary materials.}




    
    


    


    



    
    

    


    
    We should also note that out-of-domain training exhibits varying performance depending on the specific training--test dataset pairs. Training with datasets that are similar tends to improve model accuracy. For instance, the best out-of-domain results for FMT-C and FMT-M are achieved when each is trained on the other, as they share both visual and semantic characteristics that make them more compatible than other sources. Conversely, training with a synthetic dataset does not yield better results, even though it contains a larger number of samples. This suggests that the similarity between training and test data plays a more critical role in performance than the sheer quantity of training material.

    Lastly, it is worth highlighting that the end-to-end pipeline performs best in most out-of-domain scenarios, even though in some cases the best F1\textsubscript{m} score is near random. 


    In the next subsections we will examine in detail the results produced by each of the studied methods.

    \subsection{CRRN for transcription results}

    \begin{table*}[tbp]
 \centering
 \resizebox{\linewidth}{!}{
 \begin{tabular}{ll|rrr|rrrrrrr}
 \hline
 \textbf{Training} & \textbf{Target} & & & & \multicolumn{7}{c}{\textbf{F1 per class}} \\
 \textbf{dataset} & \textbf{dataset} & \multirow[b]{-2}{*}{\textbf{SER}} & \multirow[b]{-2}{*}{\textbf{F1\textsubscript{M}}} & \multirow[b]{-2}{*}{\textbf{F1\textsubscript{m}}} & \multicolumn{1}{c}{\textbf{Clef}} & \multicolumn{1}{c}{\textbf{Meter}} & \multicolumn{1}{c}{\textbf{Pitch}} & \multicolumn{1}{c}{\textbf{Pitch SI}} & \multicolumn{1}{c}{\textbf{Interval}} & \multicolumn{1}{c}{\textbf{Rhythm}} & \multicolumn{1}{c}{\textbf{Melody}} \\ \hline
 & FMT-C & \cellcolor[HTML]{EFEFEF}15.57 & \cellcolor[HTML]{EFEFEF}73.8 & \cellcolor[HTML]{EFEFEF}75.2 & \cellcolor[HTML]{EFEFEF}- & \cellcolor[HTML]{EFEFEF}64.6 & \cellcolor[HTML]{EFEFEF}76.8 & \cellcolor[HTML]{EFEFEF}79.5 & \cellcolor[HTML]{EFEFEF}74.9 & \cellcolor[HTML]{EFEFEF}80.0 & \cellcolor[HTML]{EFEFEF}\underline{67.2} \\
 & FMT-M & 72.04 & 11.8 & 9.6 & - & 59.1 & 0.5 & 0.5 & 1.0 & 10.1 & 0.0 \\
 & Malaga & 78.02 & 5.2 & 2.1 & 10.0 & 26.5 & 0.0 & 0.0 & 0.0 & 0.0 & 0.0 \\
 \multirow{-4}{*}{\textbf{FMT-C}} & PrIMuS & 89.54 & 7.8 & 5.4 & 45.0 & 7.6 & 0.0 & 0.1 & 0.1 & 1.6 & 0.0 \\ \hline
 & FMT-C & 84.83 & 5.1 & 3.3 & - & 27.3 & 1.2 & 0.0 & 0.0 & 2.4 & 0.0 \\
 & FMT-M & \cellcolor[HTML]{EFEFEF}11.27 & \cellcolor[HTML]{EFEFEF}85.9 & \cellcolor[HTML]{EFEFEF}86.1 & \cellcolor[HTML]{EFEFEF}- & \cellcolor[HTML]{EFEFEF}85.1 & \cellcolor[HTML]{EFEFEF}87.6 & \cellcolor[HTML]{EFEFEF}87.7 & \cellcolor[HTML]{EFEFEF}86.3 & \cellcolor[HTML]{EFEFEF}\underline{88.1} & \cellcolor[HTML]{EFEFEF}80.5 \\
 & Malaga & 92.95 & 7.2 & 3.6 & 10.0 & 40.0 & 0.0 & 0.0 & 0.0 & 0.8 & 0.0 \\
 \multirow{-4}{*}{\textbf{FMT-M}} & PrIMuS & 92.86 & 8.3 & 5.7 & 45.0 & 12.9 & 0.0 & 0.0 & 0.0 & 0.0 & 0.0 \\ \hline
 & FMT-C & 83.78 & 2.9 & 2.2 & - & 10.0 & 0.0 & 0.0 & 0.0 & 7.6 & 0.0 \\
 & FMT-M & 89.92 & 10.0 & 7.8 & - & 43.6 & 0.0 & 0.5 & 0.0 & 16.0 & 0.0 \\
 & Malaga & \cellcolor[HTML]{EFEFEF}27.00 & \cellcolor[HTML]{EFEFEF}51.6 & \cellcolor[HTML]{EFEFEF}49.9 & \cellcolor[HTML]{EFEFEF}44.8 & \cellcolor[HTML]{EFEFEF}84.8 & \cellcolor[HTML]{EFEFEF}42.2 & \cellcolor[HTML]{EFEFEF}48.1 & \cellcolor[HTML]{EFEFEF}39.1 & \cellcolor[HTML]{EFEFEF}66.1 & \cellcolor[HTML]{EFEFEF}36.4 \\
 \multirow{-4}{*}{\textbf{Malaga}} & PrIMuS & 95.90 & 9.9 & 7.2 & 41.6 & 27.6 & 0.0 & 0.0 & 0.0 & 0.0 & 0.0 \\ \hline
 & FMT-C & 131.65 & 1.1 & 1.4 & - & 0.0 & 0.0 & 0.0 & 0.0 & 6.7 & 0.0 \\
 & FMT-M & 82.03 & 1.0 & 1.0 & - & 1.4 & 0.0 & 0.0 & 0.0 & 4.9 & 0.0 \\
 & Malaga & 126.48 & 4.8 & 1.5 & 24.0 & 5.4 & 0.0 & 0.0 & 0.0 & 4.0 & 0.0 \\
 \multirow{-4}{*}{\textbf{PrIMuS}} & PrIMuS & \cellcolor[HTML]{EFEFEF}\underline{3.34} & \cellcolor[HTML]{EFEFEF}\underline{98.1} & \cellcolor[HTML]{EFEFEF}\underline{97.7} & \cellcolor[HTML]{EFEFEF} \underline{99.9} & \cellcolor[HTML]{EFEFEF} \underline{99.8} & \cellcolor[HTML]{EFEFEF}\underline{97.2} & \cellcolor[HTML]{EFEFEF}\underline{97.3} & \cellcolor[HTML]{EFEFEF}\underline{97.2} & \cellcolor[HTML]{EFEFEF}\underline{98.2} & \cellcolor[HTML]{EFEFEF}\underline{96.8} \\ \hline
 & FMT-C & \underline{14.61}\cellcolor[HTML]{EFEFEF} & \underline{76.3}\cellcolor[HTML]{EFEFEF} & \underline{77.6}\cellcolor[HTML]{EFEFEF} & -\cellcolor[HTML]{EFEFEF} & \underline{68.8}\cellcolor[HTML]{EFEFEF} & \underline{80.4}\cellcolor[HTML]{EFEFEF} & \underline{82.2}\cellcolor[HTML]{EFEFEF} & \underline{78.2}\cellcolor[HTML]{EFEFEF} & \underline{81.7}\cellcolor[HTML]{EFEFEF} & 66.7\cellcolor[HTML]{EFEFEF} \\
 & FMT-M & \underline{10.14}\cellcolor[HTML]{EFEFEF} & \underline{88.5}\cellcolor[HTML]{EFEFEF} & \underline{88.2}\cellcolor[HTML]{EFEFEF} & -\cellcolor[HTML]{EFEFEF} & \underline{92.4}\cellcolor[HTML]{EFEFEF} & \underline{89.8}\cellcolor[HTML]{EFEFEF} & \underline{90.7}\cellcolor[HTML]{EFEFEF} & \underline{88.7}\cellcolor[HTML]{EFEFEF} & 87.9\cellcolor[HTML]{EFEFEF} & \underline{81.6}\cellcolor[HTML]{EFEFEF} \\
 & Malaga & \underline{23.63}\cellcolor[HTML]{EFEFEF} & \underline{59.3}\cellcolor[HTML]{EFEFEF} & \underline{58.7}\cellcolor[HTML]{EFEFEF} & \underline{48.1}\cellcolor[HTML]{EFEFEF} & \underline{85.7}\cellcolor[HTML]{EFEFEF} & \underline{54.0}\cellcolor[HTML]{EFEFEF} & \underline{57.5}\cellcolor[HTML]{EFEFEF} & \underline{48.6}\cellcolor[HTML]{EFEFEF} & \underline{73.8}\cellcolor[HTML]{EFEFEF} & \underline{47.2}\cellcolor[HTML]{EFEFEF} \\
 \multirow{-4}{*}{\textbf{MultiDomain}} & PrIMuS & 91.41 & 11.7 & 7.9 & 59.9 & 19.9 & 0.1 & 0.1 & 0.1 & 2.0 & 0 \\
 \hline
 \end{tabular}
 }
 \caption{Results of the transcription-based approach. Transcription quality is evaluated using SER, while retrieval over the transcriptions is measured with F1. Macro (F1\textsubscript{M}) and micro (F1\textsubscript{m}) F1 scores provide an overall summary across all query types, and per-class F1 values show results for individual query categories. In cases where clef-related queries were discarded, they were excluded from the computation of the overall F1 metrics. In-domain cases are highlighted in gray, and top scores for each dataset are underlined.}
 \label{tab:transcription}
\end{table*}

        We now evaluate the transcription quality and its impact on query retrieval across different training scenarios (Table~\ref{tab:transcription}).
 
        As expected, in-domain transcription achieves low SER values, depending on the dataset size and difficulty, resulting in generally accurate query retrieval. The multi-domain approach further improves these in-domain results, but it is arguably as unrealistic as in-domain training in practical scenarios, since obtaining a comprehensive multi-domain training partition can be very costly. A more realistic study could involve a multi-domain approach with a reduced number of in-domain samples, reflecting cases where data availability is limited.


         In contrast, out-of-domain experiments show virtually no generalization: the transcription exhibits very high SER, and consequently, almost no positive queries are retrieved, demonstrating that poor transcription quality directly leads to very poor query retrieval. Across all real datasets, only the meter and clef queries are occasionally retrieved correctly. This suggests that, even when other aspects fail, the model reliably learns to encode the kern header fields \texttt{(**kern, *clefG2, <meter>)}. This hypothesis is supported by the observation that training on either FMT partition yields identical clef F1 values for the Malaga and PrIMuS datasets. Given that all FMT pieces start in clef G, this likely explains the results.

         A notable exception to this trend is that the rhythm retrieval metric for OOD experiments is not always zero. This indicates that note duration is the only characteristic beyond the score header that the model is able to generalize, albeit only slightly. For all other query types, the results suggest that the model fails to capture any meaningful information that could be leveraged in out-of-domain scenarios.

\begin{table*}[tbp]
 \centering
 \resizebox{\linewidth}{!}{
\begin{tabular}{ll|rr|rrrrrrr}
\hline
\textbf{Training} & \textbf{Target} & & & \multicolumn{7}{c}{\textbf{F1 per class}} \\
\textbf{dataset} & \textbf{dataset} & \multirow[b]{-2}{*}{\textbf{F1\textsubscript{M}}} & \multirow[b]{-2}{*}{\textbf{F1\textsubscript{m}}} & \textbf{Clef} & \textbf{Meter} & \textbf{Pitch} & \textbf{Pitch SI} & \textbf{Interval} & \textbf{Rhythm} & \textbf{Melody} \\ \hline
 & FMT-C & \cellcolor[HTML]{EFEFEF}\underline{52.9} & \cellcolor[HTML]{EFEFEF}49.3 & \cellcolor[HTML]{EFEFEF}- & \cellcolor[HTML]{EFEFEF}\underline{75.0} & \cellcolor[HTML]{EFEFEF}51.4 & \cellcolor[HTML]{EFEFEF}\underline{57.8} & \cellcolor[HTML]{EFEFEF}52.1 & \cellcolor[HTML]{EFEFEF}39.1 & \cellcolor[HTML]{EFEFEF}42.2 \\
 & FMT-M & 49.1 & 48.2 & - & 50.8 & 54.4 & 56.6 & 52.5 & 38.5 & 41.9 \\
 & Malaga & 41.6 & 48.1 & 0.0 & 44.0 & 58.2 & 56.7 & 54.6 & 37.9 & 39.9 \\
\multirow{-4}{*}{\textbf{FMT-C}} & PrIMuS & 34.8 & 46.3 & 0.0 & 0.0 & 53.7 & 55.1 & 54.2 & 38.6 & 42.1 \\ \hline
 & FMT-C & 51.2 & 53.8 & - & 47.1 & 54.8 & 52.8 & 54.1 & 48.0 & 50.2 \\
 & FMT-M & \cellcolor[HTML]{EFEFEF}52.8 & \cellcolor[HTML]{EFEFEF}\underline{54.0} & \cellcolor[HTML]{EFEFEF}- & \cellcolor[HTML]{EFEFEF}\underline{58.2} & \cellcolor[HTML]{EFEFEF}\underline{55.5} & \cellcolor[HTML]{EFEFEF}\underline{55.1} & \cellcolor[HTML]{EFEFEF}\underline{53.7} & \cellcolor[HTML]{EFEFEF}47.3 & \cellcolor[HTML]{EFEFEF}47.1 \\
 & Malaga & 42.8 & 50.4 & 0.0 & 50.9 & 55.3 & 55.0 & 53.9 & 42.7 & 42.3 \\
\multirow{-4}{*}{\textbf{FMT-M}} & PrIMuS & 42.1 & 50.3 & 0.0 & 39.6 & 54.3 & 53.8 & 53.9 & 47.4 & 45.6 \\ \hline
 & FMT-C & 48.1 & 52.1 & - & 34.1 & 53.9 & 49.7 & 55.2 & 47.7 & 47.8 \\
 & FMT-M & 53.4 & 54.6 & - & 51.1 & 55.8 & 54.7 & 53.0 & 51.4 & 54.3 \\
 & Malaga & \cellcolor[HTML]{EFEFEF}\underline{45.4} & \cellcolor[HTML]{EFEFEF}\underline{54.2} & \cellcolor[HTML]{EFEFEF}0.0 & \cellcolor[HTML]{EFEFEF}\underline{51.8} & \cellcolor[HTML]{EFEFEF}\underline{61.0} & \cellcolor[HTML]{EFEFEF}50.8 & \cellcolor[HTML]{EFEFEF}53.6 & \cellcolor[HTML]{EFEFEF}\underline{47.0} & \cellcolor[HTML]{EFEFEF}\underline{53.9} \\
\multirow{-4}{*}{\textbf{Malaga}} & PrIMuS & 51.0 & 52.6 & 48.5 & 52.3 & 54.3 & 53.2 & 53.1 & 47.7 & 48.4 \\ \hline
 & FMT-C & 51.6 & 52.6 & - & 57.4 & 57.8 & 53.5 & 53.4 & 39.2 & 48.3 \\
 & FMT-M & 50.1 & 49.9 & - & 57.7 & 52.0 & 52.6 & 51.1 & 39.0 & 48.4 \\
 & Malaga & 43.2 & 56.5 & 32.6 & 51.8 & 59.1 & 57.6 & 53.0 & 48.6 & 0.0 \\
\multirow{-4}{*}{\textbf{PrIMuS}} & PrIMuS & \cellcolor[HTML]{EFEFEF}\underline{74.9} & \cellcolor[HTML]{EFEFEF}\underline{74.6} & \cellcolor[HTML]{EFEFEF}\underline{97.0} & \cellcolor[HTML]{EFEFEF}\underline{85.9} & \cellcolor[HTML]{EFEFEF}\underline{83.1} & \cellcolor[HTML]{EFEFEF}\underline{76.3} & \cellcolor[HTML]{EFEFEF}\underline{67.2} & \cellcolor[HTML]{EFEFEF}\underline{66.4} & \cellcolor[HTML]{EFEFEF}48.2 \\ \hline
 & FMT-C & 51.6\cellcolor[HTML]{EFEFEF} & \underline{53.3}\cellcolor[HTML]{EFEFEF} & -\cellcolor[HTML]{EFEFEF} & 44.9\cellcolor[HTML]{EFEFEF} & \underline{54.4}\cellcolor[HTML]{EFEFEF} & 51.9\cellcolor[HTML]{EFEFEF} & \underline{54.4}\cellcolor[HTML]{EFEFEF} & \underline{53.0}\cellcolor[HTML]{EFEFEF} & \underline{50.9}\cellcolor[HTML]{EFEFEF} \\
 & FMT-M & \underline{53.1}\cellcolor[HTML]{EFEFEF} & 53.3\cellcolor[HTML]{EFEFEF} & -\cellcolor[HTML]{EFEFEF} & 54.5\cellcolor[HTML]{EFEFEF} & 54.2\cellcolor[HTML]{EFEFEF} & 53.6\cellcolor[HTML]{EFEFEF} & 53.4\cellcolor[HTML]{EFEFEF} & \underline{50.7}\cellcolor[HTML]{EFEFEF} & \underline{52.2}\cellcolor[HTML]{EFEFEF} \\
 & Malaga & 44.2\cellcolor[HTML]{EFEFEF} & \underline{54.2}\cellcolor[HTML]{EFEFEF} & 0.0\cellcolor[HTML]{EFEFEF} & 47.3\cellcolor[HTML]{EFEFEF} & 52.5\cellcolor[HTML]{EFEFEF} & \underline{54.7}\cellcolor[HTML]{EFEFEF} & \underline{55.6}\cellcolor[HTML]{EFEFEF} & 46.8\cellcolor[HTML]{EFEFEF} & 52.7\cellcolor[HTML]{EFEFEF} \\
\multirow{-4}{*}{\textbf{MultiDomain}} & PrIMuS & 47.7 & 54.9 & 0.0 & 63.6 & 55.2 & 54.3 & 54.3 & 52.5 & \underline{53.8} \\
\hline
\end{tabular}
 }
 \caption{Results of the end-to-end approach. This table follows the same conventions as Table~\ref{tab:transcription}, with the omission of the SER column. The results correspond to the best-performing combination of parameters for each training dataset, specified as (filter size, number of decoder layers): FMT-C (128,2), FMT-M (128,1), Malaga (64,1), PrIMuS (64,1), and MultiDomain (64,2).}
 \label{tab:e2e-crossdomain}
\end{table*}


    \subsection{End-to-end Transformer results}

        A general look at the end-to-end results in Table~\ref{tab:e2e-crossdomain} again confirms that in-domain experimentation achieves the best performance. However, this approach does not benefit substantially from additional diverse samples, as evidenced by the lower results under multi-domain training.

        Moreover, the strong results obtained for ID PrIMuS experiments demonstrate that this methodology---bypassing symbolic encoding of the score images entirely---is indeed feasible. The poorer performance on other datasets may reflect the data-hungry nature of Transformer-based models, as only the synthetic dataset provides enough samples to enable effective training. However, our goal is to develop a realistic approach for query search, and in practice such large-scale datasets are rarely available. This highlights a key limitation: while end-to-end methods trained in-domain can work smoothly in data-rich scenarios, their applicability in real-world settings remains constrained by the scarcity of training material.

        To analyze the out-of-domain results, it is important to recall that a random binary classifier would achieve an F1\textsubscript{m}, or individual F1 per class, of approximately 44\%. With this baseline in mind, we observe that OOD experiments generally perform better than random, with the notable exception of rhythm queries. This may be because the model's visual focus allows it to consistently detect the head-note positions on the staff, while struggling with the varying visual representations of figure values. Moreover, although not achieving the best overall results, training with the PrIMuS dataset yields performance comparable to the best OOD scores obtained for any real dataset, suggesting that synthetic data can still provide a useful learning signal for visual approaches.

        Overall, these results indicate that, although end-to-end architectures still face significant challenges in generalization, they outperform the near-null results of the transcription-based methodology in out-of-domain settings. This suggests that bypassing symbolic transcription can provide a meaningful advantage, even when only limited data is available, highlighting the potential of end-to-end approaches for realistic query searches.


\begin{table*}[tbp]
\centering
\resizebox{\textwidth}{!}{%
\begin{tabular}{ll|rr|rrrrrrrr}
\hline
\textbf{Training} & \textbf{Target} & & & \multicolumn{7}{c}{\textbf{F1 per class}} \\
\textbf{dataset} & \textbf{dataset} & \multirow[b]{-2}{*}{\textbf{F1\textsubscript{M}}} & \multirow[b]{-2}{*}{\textbf{F1\textsubscript{m}}} & \multicolumn{1}{c}{\textbf{Clef}} & \multicolumn{1}{c}{\textbf{Meter}} & \multicolumn{1}{c}{\textbf{Pitch}} & \multicolumn{1}{c}{\textbf{Pitch SI}} & \multicolumn{1}{c}{\textbf{Interval}} & \multicolumn{1}{c}{\textbf{Rhythm}} & \multicolumn{1}{c}{\textbf{Melody}} \\ \hline
 & FMT-C & \cellcolor[HTML]{EFEFEF}\underline{51.6} & \cellcolor[HTML]{EFEFEF}\underline{57.9} & \cellcolor[HTML]{EFEFEF}- & \cellcolor[HTML]{EFEFEF}\underline{78.3} & \cellcolor[HTML]{EFEFEF}\underline{62.1} & \cellcolor[HTML]{EFEFEF}\underline{56.6} & \cellcolor[HTML]{EFEFEF}52.3 & \cellcolor[HTML]{EFEFEF}\underline{51.8} & \cellcolor[HTML]{EFEFEF}\underline{60.4} \\
 & FMT-M & 25.2 & 29.3 & - & 40.5 & 28.4 & 16.4 & 42.5 & 28.2 & 20.4 \\
 & Malaga & 22.1 & 28.2 & 0.0 & 26.0 & 19.1 & 12.8 & 46.9 & 37.6 & 12.2 \\
\multirow{-4}{*}{\textbf{FMT-C}} & PrIMuS & 27.5 & 26.6 & 33.3 & 42.1 & 19.8 & 17.5 & 35.6 & 39.8 & 14.5 \\ \hline
 & FMT-C & 31.1 & 37.9 & - & 29.5 & 33.5 & 24.7 & 35.2 & 50.9 & 43.7 \\
 & FMT-M & \cellcolor[HTML]{EFEFEF}\underline{50.6} & \cellcolor[HTML]{EFEFEF}\underline{55.6} & \cellcolor[HTML]{EFEFEF}- & \cellcolor[HTML]{EFEFEF}\underline{90.4} & \cellcolor[HTML]{EFEFEF}\underline{49.0} & \cellcolor[HTML]{EFEFEF}49.9 & \cellcolor[HTML]{EFEFEF}\underline{47.8} & \cellcolor[HTML]{EFEFEF}\underline{59.9} & \cellcolor[HTML]{EFEFEF}\underline{57.1} \\
 & Malaga & 33.1 & 39.6 & 0.0 & 38.9 & 28.3 & 24.3 & 49.9 & 46.7 & 43.5 \\
\multirow{-4}{*}{\textbf{FMT-M}} & PrIMuS & 33.1 & 40.2 & 0.0 & 19.4 & 35.3 & 33.0 & 41.8 & 54.6 & 47.9 \\ \hline
 & FMT-C & 34.3 & 45.4 & - & 8.0 & 52.4 & 49.2 & 47.1 & 38.0 & 45.1 \\
 & FMT-M & 32.3 & 41.0 & - & 16.9 & 49.8 & 42.6 & 44.0 & 35.6 & 37.2 \\
 & Malaga & \cellcolor[HTML]{EFEFEF}\underline{46.1} & \cellcolor[HTML]{EFEFEF}\underline{47.7} & \cellcolor[HTML]{EFEFEF}\underline{10.5} & \cellcolor[HTML]{EFEFEF}\underline{85.7} & \cellcolor[HTML]{EFEFEF}\underline{36.8} & \cellcolor[HTML]{EFEFEF}\underline{37.5} & \cellcolor[HTML]{EFEFEF}\underline{37.8} & \cellcolor[HTML]{EFEFEF}\underline{55.8} & \cellcolor[HTML]{EFEFEF}\underline{58.3} \\
\multirow{-4}{*}{\textbf{Malaga}} & PrIMuS & 37.5 & 39.8 & 04.2 & 76.3 & 41.4 & 40.1 & 36.3 & 34.0 & 29.8 \\ \hline
 & FMT-C & 30.8 & 42.8 & - & 16.9 & 49.5 & 41.7 & 1.2 & 51.3 & 55.1 \\
 & FMT-M & 30.7 & 38.7 & - & 28.0 & 43.3 & 45.6 & 11.2 & 40.9 & 45.5 \\
 & Malaga & 36.4 & 37.8 & 33.7 & 49.1 & 39.4 & 30.1 & 7.5 & 46.9 & 48.1 \\
\multirow{-4}{*}{\textbf{PrIMuS}} & PrIMuS & \cellcolor[HTML]{EFEFEF}\underline{81.6} & \cellcolor[HTML]{EFEFEF}\underline{79.1} & \cellcolor[HTML]{EFEFEF}\underline{94.5} & \cellcolor[HTML]{EFEFEF}\underline{97.0} & \cellcolor[HTML]{EFEFEF}\underline{82.7} & \cellcolor[HTML]{EFEFEF}\underline{75.0} & \cellcolor[HTML]{EFEFEF}\underline{57.2} & \cellcolor[HTML]{EFEFEF}\underline{78.3} & \cellcolor[HTML]{EFEFEF}\underline{86.5} \\ \hline
 & FMT-C & 45.9\cellcolor[HTML]{EFEFEF} & 50.7\cellcolor[HTML]{EFEFEF} & -\cellcolor[HTML]{EFEFEF} & 76.0\cellcolor[HTML]{EFEFEF} & 50.0\cellcolor[HTML]{EFEFEF} & 48.8\cellcolor[HTML]{EFEFEF} & \underline{53.1}\cellcolor[HTML]{EFEFEF} & 51.3\cellcolor[HTML]{EFEFEF} & 42.2\cellcolor[HTML]{EFEFEF} \\
 & FMT-M & 43.3\cellcolor[HTML]{EFEFEF} & 47.8\cellcolor[HTML]{EFEFEF} & -\cellcolor[HTML]{EFEFEF} & 78.7\cellcolor[HTML]{EFEFEF} & 47.2\cellcolor[HTML]{EFEFEF} & \underline{52.7}\cellcolor[HTML]{EFEFEF} & 41.0\cellcolor[HTML]{EFEFEF} & 42.6\cellcolor[HTML]{EFEFEF} & 41.0\cellcolor[HTML]{EFEFEF} \\
 & Malaga & 30.1\cellcolor[HTML]{EFEFEF} & 33.1\cellcolor[HTML]{EFEFEF} & 0.0\cellcolor[HTML]{EFEFEF} & 50.4\cellcolor[HTML]{EFEFEF} & 35.4\cellcolor[HTML]{EFEFEF} & 32.8\cellcolor[HTML]{EFEFEF} & 30.0\cellcolor[HTML]{EFEFEF} & 29.9\cellcolor[HTML]{EFEFEF} & 32.5\cellcolor[HTML]{EFEFEF} \\
\multirow{-4}{*}{\textbf{MultiDomain}} & PrIMuS & 68.7 & 65.2 & 81.7 & 94.2 & 66.5 & 58.2 & 56.3 & 57.6 & 66.5 \\
\hline
\end{tabular}
}
\caption{Results for the PaliGemma 2 LLM model, finetuned for each training dataset. This table follows the same conventions as Table~\ref{tab:e2e-crossdomain}.}
\label{tab:paligemma_results}
\end{table*}

    \subsection{Multimodal LLM results}

        Lastly, we evaluate the multimodal LLM architecture PaliGemma 2 with 3B parameters. As with the other approaches, reviewing the results in Table~\ref{tab:paligemma_results} shows in-domain experimentation clearly dominates. No substantial improvements are observed through multi-domain training.
    
        The in-domain experiments generally provide results comparable to---or in some cases better than---the end-to-end pipeline, excelling on meter, rhythm, and, somewhat surprisingly due to its complexity, melody queries. This performance might stem from the model's broader musical knowledge, which allows it to visually recognize these patterns.
        
        In contrast, pitch and interval queries remain more challenging. Interval detection is inherently difficult for vision-based models, as the information is implicit 
        rather than directly observed. Similarly, non-scale pitch queries require the model to recognize that different visual representations correspond to the same pitch class. One might have expected simple pitch queries to be easier, as suggested by the end-to-end results, but the results hint otherwise. This may reflect the model's tendency to perform better on queries that involve rhythm and single symbol recognition.

        Although the model performs well on in-domain data, its out-of-domain results fall below random performance, underscoring poor generalization beyond the training distribution. This suggests that, despite leveraging some prior musical knowledge to recognize patterns, the model struggles to extract meaningful information from unfamiliar scores without sufficient domain-specific examples, reinforcing the critical role of training-test similarity in this setting.
        

\section{Conclusions}\label{conclusions}
 


    This work explored multiple methodologies for content-based search from sheet music images, ranging from traditional transcription-based pipelines to end-to-end architectures and multimodal LLMs. Each approach was evaluated across in-domain, out-of-domain, and multi-domain scenarios to assess their ability to generalize and to operate under realistic conditions.

    Although transcription-free still fall short of the transcription-based pipeline in ideal in-domain experimentation, such conditions are largely unrealistic. In practical applications, most musical collections lack symbolic transcriptions, and producing them manually remains both time-consuming and costly, making the out-of-domain setting a more realistic and meaningful evaluation scenario.

    To address data limitations, we also experimented with training on a synthetic dataset, which can be generated more easily and at larger scale. Among the methods studied, the traditional transcription-based approach failed to retain accuracy in this context, whereas the proposed end-to-end pipeline---though not outstanding---still obtained higher retrieval scores. 
    These findings suggest that visual retrieval methods hold promise for improving robustness in realistic, data-limited scenarios, paving the way for future exploration.

 
         


    Several directions remain open for future work. 
    We plan to investigate how dataset size influences retrieval quality (e.g. expanding the synthetic dataset could enhance generalization) or whether combining a multi-domain dataset with only a small number of in-domain samples could yield comparable results on real data.

    In terms of query design, future research will focus on finding an accurate key detection method to enable degree-based queries, as well as extending the system's expressiveness by incorporating partial matching and wildcard tokens to better reflect realistic search behaviors.

\section*{Declarations}


\subsection{Funding}
This research was supported by the Spanish Ministry of Science and Innovation through the LEMUR research project (PID2023-148259NB-I00), funded by MCIU/AEI/10.13039/501100011033/FEDER, EU, and the European Social Fund Plus (FSE+).

\subsection{Competing interests}
The authors have no relevant financial or non-financial interests to disclose.









\bibliography{sn-bibliography}

\end{document}